\documentclass[conference]{IEEEtran}
\IEEEoverridecommandlockouts
\usepackage{cite}
\usepackage{amsmath,amssymb,amsfonts}
\usepackage{algorithmic}
\usepackage{graphicx}
\usepackage{textcomp}
\usepackage{xcolor}
\usepackage{pgfplots}
\def\BibTeX{{\rm B\kern-.05em{\sc i\kern-.025em b}\kern-.08em
    T\kern-.1667em\lower.7ex\hbox{E}\kern-.125emX}}
\begin{document}

\title{{Science based AI model certification for untrained operational environments with application in traffic state estimation}\\
}

\author{\IEEEauthorblockN{Daryl Mupupuni\IEEEauthorrefmark{1},
Anupama Guntu\IEEEauthorrefmark{2}, Liang Hong\IEEEauthorrefmark{3},
Kamrul Hasan\IEEEauthorrefmark{4} and
Leehyun Keel\IEEEauthorrefmark{5}}
\IEEEauthorblockA{Department of Electrical and Computer Engineering,
Tennessee State University\\
Nashville, TN 37209, USA\\
Email: \IEEEauthorrefmark{1}darylmupupuni@gmail.com,
\IEEEauthorrefmark{2}anupamaguntu23@gmail.com,
\IEEEauthorrefmark{3}lhong@tnstate.edu, \\
\IEEEauthorrefmark{4}mhasan1@tnstate.edu,
\IEEEauthorrefmark{5}keel.tsu@gmail.com}}

\maketitle

\begin{abstract}
The expanding role of Artificial Intelligence (AI) in diverse engineering domains highlights the challenges associated with deploying AI models in new operational environments, involving substantial investments in data collection and model training. Rapid application of AI necessitates evaluating the feasibility of utilizing pre-trained models in unobserved operational settings with minimal or no additional data. However, interpreting the opaque nature of AI's black-box models remains a persistent challenge. Addressing this issue, this paper proposes a science-based certification methodology to assess the viability of employing pre-trained data-driven models in untrained operational environments. The methodology advocates a profound integration of domain knowledge, leveraging theoretical and analytical models from physics and related disciplines, with data-driven AI models. This novel approach introduces tools to facilitate the development of secure engineering systems, providing decision-makers with confidence in the trustworthiness and safety of AI-based models across diverse environments characterized by limited training data and dynamic, uncertain conditions. The paper demonstrates the efficacy of this methodology in real-world safety-critical scenarios, particularly in the context of traffic state estimation. Through simulation results, the study illustrates how the proposed methodology efficiently quantifies physical inconsistencies exhibited by pre-trained AI models. By utilizing analytical models, the methodology offers a means to gauge the applicability of pre-trained AI models in new operational environments. This research contributes to advancing the understanding and deployment of AI models, offering a robust certification framework that enhances confidence in their reliability and safety across a spectrum of operational conditions.
\end{abstract}


\section{Introduction}
Artificial intelligence (AI) has experienced tremendous growth and application across several fields \cite{amin2024explainable,amin2022industrial,amin2022analysing}. Application of AI has completely changed the engineering sector over the last ten years	by allowing engineers to design creative ideas, solve complex problems, and expedite operations in several engineering disciplines. For instance, various environmental engineering problems were solved by AI based prediction models such as pollution control \cite{YE2020134279}, wastewater to improve the performance of several chemical and logical treatment processes \cite{Yetilmezsoy2011ARTIFICIALIP}. AI is also useful in autonomous systems and proactive maintenance in the aerospace and automobile industries \cite{Zhang2017CurrentTI}. AI is changing engineering by automating tasks, improving designs, and enabling predictive analysis. The integration of AI will continue driving innovation in the future \cite{xing2020driving}.

Although AI models like Artificial Neural Networks, Ensemble Approaches etc., continue to perform better than other models in many fields, but their acceptance in delicate fields like finance and healthcare is questionable because of the models’ difficulty in being understood and explained \cite{islam2019infusing}. The lack of interpretability or transparency also known as "black-box models," is a key cause of concern in some AI models and it is challenging to comprehend how they came to those conclusions \cite{zhou2017machine}. The minimal available of data is another restriction. For AI models to learn effectively, large, and varied datasets are necessary to train the models. Owing to the rarity of some events or the biases present in the data, it can be difficult to obtain balanced datasets in some areas. As a result, the model’s predictions may be biased or have skewed representation. This is also reflected in the application of models developed from these datasets to be employed in untrained operational situations \cite{7906512}.

Since the AI models have black-box nature, they need to be certified to make sure it is working efficiently in all environments. This certification can be done in several ways such as assessment certifying the fulfillment of minimum ethical requirements and what we describe as nuanced assessment \cite{article}. Some important applications like robots in work spaces shared with humans \cite{winter2021trusted}, social and environmental \cite{gupta2020secure} need to be certified. This type of certification study is also important in traffic state estimation (TSE) which is not fully examined yet. The goal of the certification process is to ascertain whether the TSE model developed under different traffic state conditions can accurately anticipate in a new setting while upholding the traffic conservation rule. By certifying the model, we hope to increase its interpretability, strengthen its dependability, and demonstrate its suitability for TSE example that occur in the real world \cite{tambon2022certify}. The findings of this study have implications for developing the TSE area and advance knowledge of certifying deep learning models for safety-critical applications. 

To fulfill the urgent demand for safety, dependability, and transparency in AI applications, deep learning models need certification \cite{winter2021trusted}. The contributions of this paper are the establishment of a thorough science-based AI model certification methodology that ensures the resilience of deep learning models in untrained operational situations by thorough examination and the incorporation of known scientific concepts. This work enables one to ascertain the extent to which AI models adapt to untrained complex and dynamic environments, reducing risks and potential accidents by bridging the gap between conventional data-driven approaches and science-based understanding. Additionally, the focus on transparency and explainability encourages a better understanding of model judgments, fostering public confidence and facilitating regulatory compliance \cite{bakirtzis2022dynamic},\cite{zhang2023ethics}. We seek to increase the precision and dependability of TSE models, resulting in safer and more effective transportation systems.

\subsection{Literature Review }

Certification of AI models is crucial when implementing them in untrained operational scenarios, especially where traditional data-driven approaches may fall short. This is particularly important in applications like autonomous vehicles, where certification ensures safety and dependability in diverse and unpredictable conditions such as bad weather and unusual traffic patterns. Similarly, in healthcare diagnostics, certification is necessary to handle different patient populations, rare diseases, and unforeseen medical situations, instilling confidence in the accuracy and clarity of AI-assisted diagnoses especially in untrained operational environments \cite{azzam2004route}, \cite{yanisky2019equality}. 

In natural disaster prediction and response artificial intelligence models are used to anticipate natural disasters, such as earthquakes, hurricanes, or wildfires, need to be certified to function well in disaster scenarios that haven’t been encountered before \cite{kuglitsch2022facilitating}. These model certifications guarantees their ability to adjust to shifting circumstances and deliver accurate early warnings. In aerospace and aviation AI certification is needed to address the difficulties of untrained operational contexts, such as air traffic changes, weather disturbances, and emergency scenarios, certification is required for autonomous drones and aircraft systems \cite{bakirtzis2022dynamic}.

In the fields of industrial robotics and manufacturing, artificial intelligence (AI) models must be certified to operate in dynamic production environments with the possibility of varying raw material supply, equipment breakdowns, or unforeseen production difficulties.AI models used for environmental monitoring, such as those that predict weather, ocean currents, or air quality, must be certified to handle a variety of geographic and climatic conditions \cite{SOORI202354}. 

Certifying AI models is paramount in security and surveillance for operation in uncontrolled environments, such as public spaces where unforeseen occurrences and abnormalities may occur \cite{berghoff2020vulnerabilities}. Financial risk assessment also needs certification owing to financial markets being subject to abrupt volatility and unforeseen events, AI models for risk assessment and prediction in finance must be approved for untrained operational settings. For Power Grid Management, AI models used to manage power grids and energy distribution systems must be certified to react to unforeseen changes in energy supply, demand, and potential grid failures \cite{syu2023ai}. For application in space exploration, AI models used in space missions need to be certified to withstand the harsh environments of uncharted space, where data may be sporadic and unexpected \cite{budnik2018guided}. 

Certification is necessary in order to check for several challenges that are encountered when we are trying to employ pre-trained AI models in new environments. Some of the challenges that may be encountered include models that may be vulnerable to adversarial assaults, in which malicious inputs are skillfully created to falsely estimate outcomes \cite{berghoff2020vulnerabilities}. When pre-trained models are used in untrained environments, privacy concerns may arise because vulnerable data may unintentionally leak through model outputs. Ethics issues may also arise if AI models are used in untrained environments without being properly vetted or knowing their limitations, especially if important choices are based on their outputs. Pre-trained models can be used in situations where they are not in compliance with rules or legal requirements, which can result in legal issues and liability issues \cite{hussain2022shape}. Understanding how pre-trained models work can be tricky because they are complex sometimes and difficult to interpret. 

Another important application is traffic state estimation where this certification is mandatory to test the models in different environments.  It is always difficult to acquire huge number of data in all environments since it involves more cost to collect it. In that scenario, certification of AI model in other environments is necessary.  Thus, it is essential to carry out thorough testing, validation, and fine-tuning of the pre-trained model in the particular operating context in order to assure its safety, dependability, and ethical use in order to reduce these risks. Additionally, keeping an eye on the model’s performance and user input can aid in spotting and resolving possible problems \cite{berghoff2020vulnerabilities}.

Pre-trained models can be quite effective and helpful in terms of time and computational power, but they may not always function at their best in novel and uncharted circumstances thus pausing risks of increased uncertainties. Pre-trained models that are typically trained on a small number of datasets may not have a thorough understanding of domains that are unrelated to their training data. The availability of unbalanced datasets is a frequent problem in traffic state estimation. A dataset that is uneven in the context of the traffic state estimation example would be one where some traffic states or situations are over represented while others are underrepresented or rare \cite{liang2018deep}. The imbalance may be caused numerous things, including traffic patterns, variable levels of congestion, or the scarcity of data from particular places or times. For instance the traffic dataset that may be readily available is mainly the NGSIM-CA data from California and the traffic state may not fully represent the state of traffic in other states say Tennessee.  The model may generate inaccurate or deceptive results in untrained circumstances as a result of its ignorance of the relevant domain. In some cases the input data in a context for which the model has not yet been trained differs from that context, the model may interpret the input incorrectly and produce false results.

Pre-trained models may inherit biases from their training data, which is known as bias amplification. These biases may intensify when applied in unknown environments, producing unfair or discriminating outcomes. It is possible for pre-trained models to over-fit to specific applications or datasets \cite{zhou2020sentix}. They might not generalize when applied in novel and varied circumstances.  Due to domain variations, the model's performance may suffer in untrained circumstances, producing inconsistent results and decreasing user confidence.

It can be very beneficial to use pre-trained AI models in untrained environments, but doing so requires careful preparation, ongoing monitoring, and the right tweaks to make sure the model operates properly and safely. Striking a balance between using the model's pre-trained skills and tailoring it to the unique demands of the operational setting is crucial.

\subsection{Methodology}

This section of the paper provides background information on traffic flow physics and its relevance to AI models, particularly in the context of traffic data certification. It emphasizes the importance of understanding fundamental traffic concepts, illustrated by the fundamental traffic diagram (Figure 1), which depicts the nonlinear relationship between traffic speed, flow, and density \cite{wang2020estimating,RAISSI2019686,lv2014traffic}. The conservation of vehicles principle, based on mass conservation, is introduced as a fundamental physics concept governing traffic flow. Additionally, mathematical models such as the Lighthill-Whitham-Richards (LWR) model, based on fluid dynamics principles, are discussed for representing traffic flow and analyzing congestion \cite{raissi2018deep,10105558}.

\begin{figure}
	\centering 
	\includegraphics[width=0.5\textwidth, angle=0]{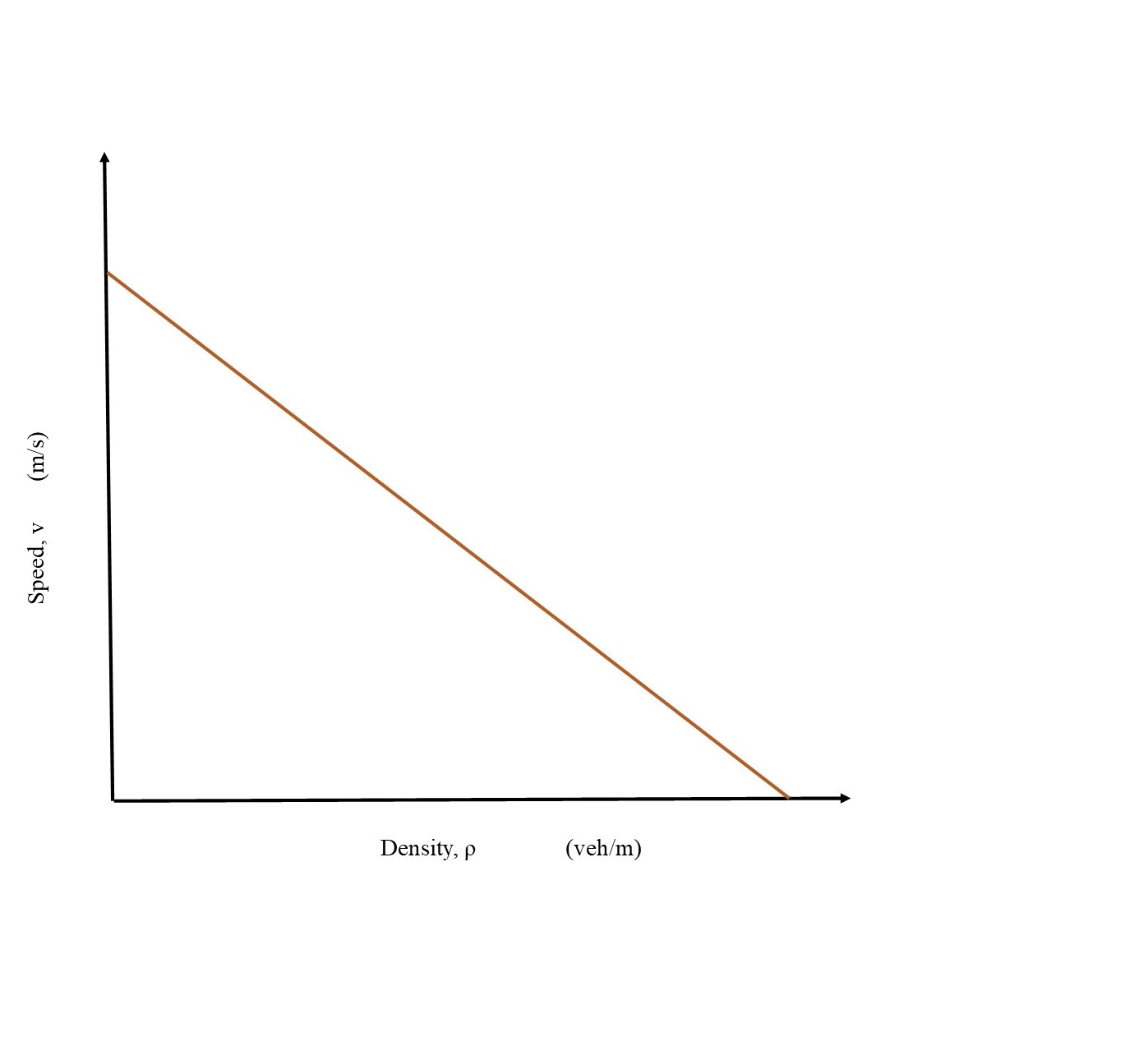}	
	\caption{Speed against density diagram in traffic flow } 
	\label{fig_mom0}%
\end{figure}

The section delves into the relationship between traffic density, flow rate, and vehicle speed, highlighting the impact of these factors on traffic conditions \cite{park2016real}. It also introduces equations describing traffic flow, including the LWR equation. The Greenshields' fundamental diagram is presented to summarize the relationship between traffic density, flow rate, and free-flow speed is mathematically defined by equations 1 and 2 \cite{greenshields1935study,XING2022127079}.

\begin{equation}
q(x,t) = \frac{\partial N(x,t)}{\partial t}
\end{equation}

\begin{equation}
\rho(x,t) = -\frac{\partial N(x,t)}{\partial x}
\end{equation}

The discussion transitions to the application of Artificial Intelligence (AI) in traffic state estimation, categorizing it as a data-driven approach. The potential of AI algorithms, particularly deep learning models, in forecasting traffic conditions is emphasized, along with the importance of labeled training data and computational resources \cite{9744160,8694956}. The structure of deep learning neural networks and the role of the cost function in training are explained. The paper provides equations for the mean squared error cost function used in traffic state estimation for the assessment of the metrics.The cost function, also known as the loss or objective function, is a crucial mathematical measure in deep learning that quantifies the disparity between a neural network's expected output and the actual target values \cite{10105558}.

Accuracy in deep learning is explored, focusing on classification and regression accuracy calculations and is defined by the equation. 

\begin{equation}
    A_{D_L} = \frac{\sqrt{\sum_{x=1}^{X_m}\sum_{t=1}^{T_n}|\rho(x,t)- \hat{\rho}(x,t)|^2}}{\sqrt{\sum_{x=1}^{X_m}\sum_{t=1}^{T_n}|\rho(x,t)|^2}}
\end{equation}

Equation 3 is used to evaluate the performance of an AI model and the results obtained from this equation are the ones that plot figure 5. 

Regression accuracy is specifically discussed in the context of traffic state estimation, introducing the Root Mean Squared Error (RMSE) formula. 

\begin{equation}
    C_{D_L} = \text{MSE}_{(\rho(x,t),\hat{\rho}(x,t))} = \frac{1}{N} \sum_{i=1}^{N} |\rho(x,t) - \hat{\rho}(x,t)|^2
\end{equation}

Equation 4 is used in the training process. N is the number of estimating outputs in Mean Squared Error MSE. The estimated vehicle density $\hat{\rho}(x,t)$ and the actual vehicle density, $\rho(x,t)$ at location x and time t.

\section{Methodology Discussion: Model Assessment against Science Metrics Laws}
To evaluate and certify AI models, the steps shown in figure 2 are followed.

Certifying the suitability of a Traffic State Estimation AI model for deployment in new environments necessitates the integration of scientific principles and traffic laws into the model assessment. This involves a comprehensive examination of the model's predictions in alignment with established scientific principles, with a primary focus on the implementation of the conservation of traffic. This approach serves to foster the development of dependable models that strictly adhere to physical and mathematical constraints, thereby establishing a foundational framework for the scientific validation and acceptance of the AI model. Upon saving the model parameters, a crucial step ensues, where the models undergo comparison with well-known physical constraints and physics laws. This comparative analysis aims to augment the reliability, robustness, and trustworthiness of deep learning models designed for traffic state estimation.

The pivotal role played by the law of conservation of traffic in this process is evident as predictions from the model are scrutinized for deviations from ideality. Ensuring adherence to fundamental principles involves applying traffic flow conservation rules or laws, validating that the predicted values for density, velocity, and predicted flow rate align with the actual influx and outflow of vehicles. The assessment extends to confirming the consistency of predicted values for flow rate and density with the anticipated connections suggested by the traffic flow model. An analysis of results is then conducted to identify anomalies or deviations from the established rules of traffic flow, scrutinizing expected density values and traffic variables. This meticulous examination identifies areas, such as instances of congestion or unexpected correlations between flow rate and density, facilitating a comprehensive evaluation of the model's adaptability to new environments.

\section{Experimental Configuration}
Assessment of the performance of the AI system's correctness and dependability are required for science-based certification of AI models. \cite{tambon2022certify}. This section introduces a science-based certification framework for AI models, emphasizing the evaluation of assessment in untrained environments. The certification process is outlined in Figure 2.

\begin{figure}
	\centering 
	\includegraphics[width=0.5\textwidth, angle=0]     
        {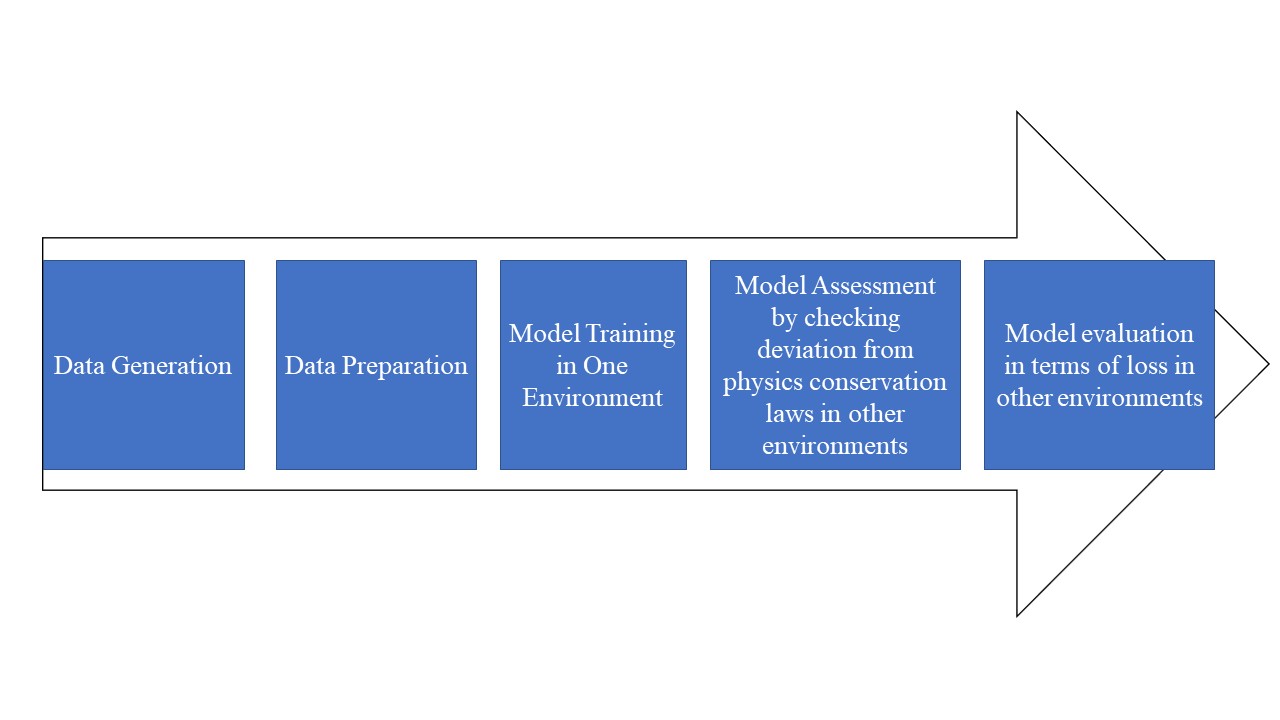}	
	\caption{Certification process} 
	\label{fig_mom0}%
\end{figure}

In order to certify AI models using physics laws for safety-critical traffic state estimation The Lax-Hopf method is used to create synthetic traffic datasets in this paper under no downstream flow and no upstream flow, and it is specifically tuned to follow the Greenshields' model and traffic conservation laws.
\begin{figure}
	\centering 
	\includegraphics[width=0.5\textwidth, angle=0]{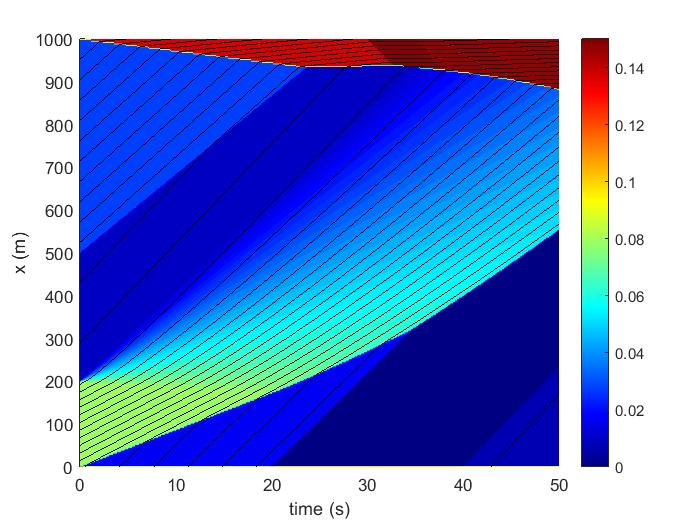}	
	\caption{Dataset generated with a $v_f$ =25} 
	\label{fig_mom0}%
\end{figure}
This experiment is done on a road segment defined by 1000 meters (x) observed over a duration of 50 seconds (t). The traffic density values throughout the domain were created by the Lax-Hopf method by setting up the following initial density values $\rho(x,t)$ and initial flow rates.

Initial density: For interval at x in  [0, 200] meters is 0.13 vehicle per meter (veh/m), [200, 500] is 0.06 veh/m and [500, 1000] is 0.03 veh/m at t=0.

Initial Flow rate: For upstream at t in [0, 20] seconds is 0.4 vehicle per second (veh/s), [20, 40] is 0.01 veh/sec and [40, 50] is 0.2 veh/sec at x=0. For downstream at t in [0, 30] seconds is 0.3 vehicle per second (veh/s), [30, 35] is 0 veh/sec and [35, 50] is 0.1 veh/sec at x=1000.

We used free flow velocity ($v_f$) is 25 meters per second (55 miles per hour) and jam density ($\rho_m$) is 0.15 veh/m. The dataset was created at the time steps of 0.1 seconds and the distance step is 2 meters. The generated synthetic dataset is displayed in figure 3.

The deep learning experimental set up uses a 10 hidden layer architecture with 40 neurons each. Two optimization algorithms were used that is the limited Broyden–Fletcher–Goldfarb–Shanno (L-BFGS) and Adam. The Tensorflow library for machine learning was implemented in the development of our Traffic State Estimation.The machine used to run the TSE was a 11th Gen Intel® Core™ i9-11900KF @ 3.50GHz × 16, with a Random Access Memory of 62.5GB. 

\section{Results and Discussions}

In conclusion, since the experiment involved first training a machine learning model to estimate traffic states, speed and flow based on sensor data. Then, using inputs road geometry and real-time traffic counts, the expected outflows were calculated by applying the physics-based conservation of traffic formula. The investigation into the comparison of various $v_f$ environments against the conservation law of traffic flow has yielded valuable insights. By generating new environments with different $v_f$ values while keeping other parameters constant, we aimed to assess the impact on the conservation law. While the conservation law provides a fundamental framework for understanding traffic flow, it is crucial to acknowledge potential deviations in real-world scenarios, influencing traffic modeling and analysis.

The estimated results from the ML model were then directly compared against the calculations derived from the established traffic flow physics model. The differences or errors between the two results were analyzed.

This allowed the paper to systematically validate the ML model's estimations and identify any discrepancies by bench marking it against well-established traffic behavior defined through fundamental physical laws. The conservation law provided a standardized way to numerically simulate "ground truth" traffic states.

Our focus on deviations from the Conservation Law in testing with different $v_f$ environments, all trained on $v_f$=25 meter/sec, is depicted in Figure 4. Notably, the analysis reveals that the lowest deviation occurs when comparing with a $v_f$=25 environment, demonstrating a consistent trend observed in Figure 5.

\begin{figure}[!ht]
  \centering
  \includegraphics[width=0.5\textwidth]{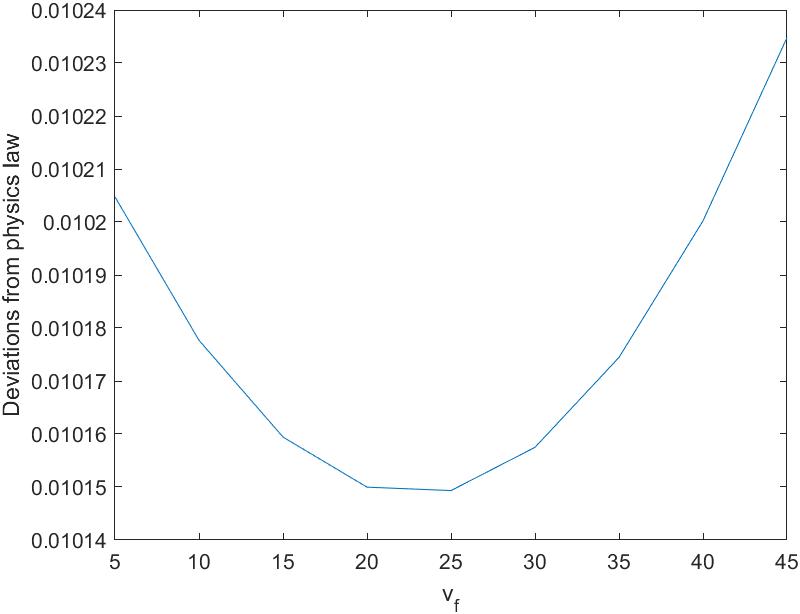}
  \caption{DL estimations for $v_f$=25.}\label{}
\end{figure}

This section further underscores the significance of certifying deep learning models with physics laws to enhance predictions in Traffic State Estimation (TSE). The deep learning model, trained in a specific environment and tested across diverse settings, particularly aligns with scenarios featuring sensors strategically positioned within the road network. This approach allows for the observation and analysis of traffic variables at fixed locations over time, mirroring real-world setups with fixed sensors.

The monitored traffic conditions and behaviors at predetermined positions contribute to predicting traffic conditions in new or unobserved areas. Through this comprehensive exploration, we emphasize the robustness of our approach, bridging the gap between deep learning models and the underlying physics governing traffic dynamics for more accurate TSE predictions. The graphical representation in Figure 4 visually reinforces the trends observed, validating the effectiveness of our methodology in real-world applications.

The deep learning (DL) model undergoes training using only 6\% of the complete dataset, specifically 15,000 samples, with a fixed $v_f$ of 25. The testing phase employs the remaining dataset, encompassing data from various environments. During training, we utilized 15,000 iterations with the Adam optimizer and an additional 50,000 iterations with the L-BFGS optimizer. The resulting DL error rates for different environments are plotted in figure 5.

As delineated in Figure 5, the lowest DL error occurs at $v_f$ = 25 m/sec, registering at 17.81 \% . This alignment with expectations is attributed to the model's training and testing being conducted within the $v_f$ = 25 m/sec environment. Notably, this finding underscores the exceptional performance of the model when trained and tested in the same environment. However, as one deviates from the conditions of the trained environment, whether in the lower or higher $v_f$ range, the DL performance diminishes, resulting in an increased DL error. This observation emphasizes the model's sensitivity to variations in the training environment, highlighting the importance of consistent conditions for optimal performance.

\begin{figure}[htbp]
  \centering
  \begin{tikzpicture}
    \begin{axis}[
        xlabel={Vf},
        ylabel={DL error},
        xmin=0, xmax=50,
        ymin=0, ymax=100,
        xtick={5, 10, 15, 20, 25, 30, 35, 40, 45},
        ytick={0, 20, 40, 60, 80, 100},
        legend pos=north east,
        grid=both,
        grid style=dashed,
      ]
      \addplot[color=blue,mark=*] coordinates {
        (5, 95.34)
        (10, 95.48)
        (15, 72.42)
        (20, 49.47)
        (25, 17.81)
        (30, 45.96)
        (35, 61)
        (40, 69.17)
        (45, 87.64)
      };
      \legend{DL error}
    \end{axis}
  \end{tikzpicture}
  \caption{DL error against Vf}
  \label{fig:dl_error}
\end{figure}
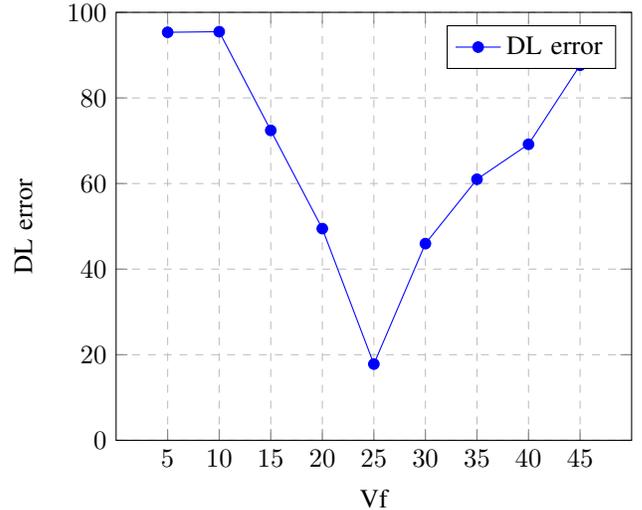


\section{Summary and Conclusions}

This paper divulged a method of science-based certification of AI models to be applied in untrained operational environments. We investigated the viability and efficacy of certifying deep learning models for safety-critical traffic state estimation using physics rules. Enhancing deep learning models' dependability and credibility in applications where precise traffic state estimation is crucial for assuring safety was the goal and data is limited. 

The findings of this work pave the door for additional approaches such as physics-regulated deep learning model and transfer learning to better employ the models in untrained operating conditions. Furthermore, in this paper we used only synthetic data, future work should utilize the real world data for betterment of certification process of the AI models.

\section*{Acknowledgements}
This work is supported in part by the National Science Foundation (NSF) under Grant No. 2130990.


\bibliographystyle{ieeetr}
\bibliography{refer}

\begin{thebibliography}{10}

\bibitem{amin2024explainable}
A.~Amin, K.~Hasan, S.~Zein-Sabatto, D.~Chimba, I.~Ahmed, and T.~Islam, ``An explainable ai framework for artificial intelligence of medical things,'' {\em arXiv preprint arXiv:2403.04130}, 2024.

\bibitem{amin2022industrial}
A.~Amin, H.~Ma, M.~S. Hossain, N.~A. Roni, E.~Haque, S.~Asaduzzaman, R.~Abedin, A.~B. Ekram, and R.~F. Akter, ``Industrial product defect detection using custom u-net,'' in {\em 2022 25th International Conference on Computer and Information Technology (ICCIT)}, pp.~442--447, IEEE, 2022.

\bibitem{amin2022analysing}
A.~Amin, H.~Ma, R.~Alam, N.~A. Roni, M.~S. Hossain, E.~Haque, A.~B. Ekram, R.~Abedin, and S.~Siddiqui, ``Analysing and detecting extreme-selfie images using ensemble technique,'' in {\em 2022 25th International Conference on Computer and Information Technology (ICCIT)}, pp.~909--914, IEEE, 2022.

\bibitem{YE2020134279}
Z.~Ye, J.~Yang, N.~Zhong, X.~Tu, J.~Jia, and J.~Wang, ``Tackling environmental challenges in pollution controls using artificial intelligence: A review,'' {\em Science of The Total Environment}, vol.~699, p.~134279, 2020.

\bibitem{Yetilmezsoy2011ARTIFICIALIP}
K.~Yetilmezsoy, B.~{\"O}zkaya, and M.~Çakmakci, ``Artificial intelligence-based prediction models for environmental engineering,'' {\em Neural Network World}, vol.~21, pp.~193--218, 2011.

\bibitem{Zhang2017CurrentTI}
T.~Zhang, Q.~Li, C.-S. Zhang, H.~Liang, P.~Li, T.~M. Wang, S.~Li, Y.~Zhu, and C.~Wu, ``Current trends in the development of intelligent unmanned autonomous systems,'' {\em Frontiers of Information Technology \& Electronic Engineering}, vol.~18, pp.~68 -- 85, 2017.

\bibitem{xing2020driving}
F.~Xing, G.~Peng, B.~Zhang, S.~Zuo, J.~Tang, and S.~Li, ``Driving innovation with the application of industrial ai in the r\&d domain,'' in {\em Distributed, Ambient and Pervasive Interactions: 8th International Conference, DAPI 2020, Held as Part of the 22nd HCI International Conference, HCII 2020, Copenhagen, Denmark, July 19--24, 2020, Proceedings 22}, pp.~244--255, Springer, 2020.

\bibitem{islam2019infusing}
S.~R. Islam, W.~Eberle, S.~Bundy, and S.~K. Ghafoor, ``Infusing domain knowledge in ai-based "black box" models for better explainability with application in bankruptcy prediction,'' 2019.

\bibitem{zhou2017machine}
L.~Zhou, S.~Pan, J.~Wang, and A.~V. Vasilakos, ``Machine learning on big data: Opportunities and challenges,'' {\em Neurocomputing}, vol.~237, pp.~350--361, 2017.

\bibitem{7906512}
A.~L’Heureux, K.~Grolinger, H.~F. Elyamany, and M.~A.~M. Capretz, ``Machine learning with big data: Challenges and approaches,'' {\em IEEE Access}, vol.~5, pp.~7776--7797, 2017.

\bibitem{article}
S.~Genovesi and J.~Mönig, ``Acknowledging sustainability in the framework of ethical certification for ai,'' {\em Sustainability}, vol.~14, p.~4157, 03 2022.

\bibitem{winter2021trusted}
P.~M. Winter, S.~Eder, J.~Weissenb{\"o}ck, C.~Schwald, T.~Doms, T.~Vogt, S.~Hochreiter, and B.~Nessler, ``Trusted artificial intelligence: Towards certification of machine learning applications,'' {\em arXiv preprint arXiv:2103.16910}, 2021.

\bibitem{gupta2020secure}
A.~Gupta, C.~Lanteigne, and S.~Kingsley, ``Secure: A social and environmental certificate for ai systems,'' 2020.

\bibitem{tambon2022certify}
F.~Tambon, G.~Laberge, L.~An, A.~Nikanjam, P.~S.~N. Mindom, Y.~Pequignot, F.~Khomh, G.~Antoniol, E.~Merlo, and F.~Laviolette, ``How to certify machine learning based safety-critical systems? a systematic literature review,'' {\em Automated Software Engineering}, vol.~29, no.~2, p.~38, 2022.

\bibitem{bakirtzis2022dynamic}
G.~Bakirtzis, S.~Carr, D.~Danks, and U.~Topcu, ``Dynamic certification for autonomous systems,'' {\em arXiv preprint arXiv:2203.10950}, 2022.

\bibitem{zhang2023ethics}
J.~Zhang and Z.-m. Zhang, ``Ethics and governance of trustworthy medical artificial intelligence,'' {\em BMC Medical Informatics and Decision Making}, vol.~23, no.~1, p.~7, 2023.

\bibitem{azzam2004route}
H.~Azzam, F.~Beaven, L.~Gill, and M.~Wallace, ``A route for qualifying/certifying an affordable structural prognostic health management (sphm) system,'' in {\em 2004 IEEE Aerospace Conference Proceedings (IEEE Cat. No. 04TH8720)}, vol.~6, pp.~3791--3808, IEEE, 2004.

\bibitem{yanisky2019equality}
S.~Yanisky-Ravid and S.~K. Hallisey, ``Equality and privacy by design: A new model of artificial intelligence data transparency via auditing, certification, and safe harbor regimes,'' {\em Fordham Urb. LJ}, vol.~46, p.~428, 2019.

\bibitem{kuglitsch2022facilitating}
M.~M. Kuglitsch, I.~Pelivan, S.~Ceola, M.~Menon, and E.~Xoplaki, ``Facilitating adoption of ai in natural disaster management through collaboration,'' {\em Nature communications}, vol.~13, no.~1, p.~1579, 2022.

\bibitem{SOORI202354}
M.~Soori, B.~Arezoo, and R.~Dastres, ``Artificial intelligence, machine learning and deep learning in advanced robotics, a review,'' {\em Cognitive Robotics}, vol.~3, pp.~54--70, 2023.

\bibitem{berghoff2020vulnerabilities}
C.~Berghoff, M.~Neu, and A.~von Twickel, ``Vulnerabilities of connectionist ai applications: evaluation and defense,'' {\em Frontiers in big Data}, vol.~3, p.~23, 2020.

\bibitem{syu2023ai}
J.-H. Syu, J.~C.-W. Lin, and G.~Srivastava, ``Ai-based electricity grid management for sustainability, reliability, and security,'' {\em IEEE Consumer Electronics Magazine}, 2023.

\bibitem{budnik2018guided}
C.~Budnik, M.~Gario, G.~Markov, and Z.~Wang, ``Guided test case generation through ai enabled output space exploration,'' in {\em Proceedings of the 13th International Workshop on Automation of Software Test}, pp.~53--56, 2018.

\bibitem{hussain2022shape}
S.~M. Hussain, D.~Buongiorno, N.~Altini, F.~Berloco, B.~Prencipe, M.~Moschetta, V.~Bevilacqua, and A.~Brunetti, ``Shape-based breast lesion classification using digital tomosynthesis images: The role of explainable artificial intelligence,'' {\em Applied Sciences}, vol.~12, no.~12, p.~6230, 2022.

\bibitem{liang2018deep}
Y.~Liang, Z.~Cui, Y.~Tian, H.~Chen, and Y.~Wang, ``A deep generative adversarial architecture for network-wide spatial-temporal traffic-state estimation,'' {\em Transportation Research Record}, vol.~2672, no.~45, pp.~87--105, 2018.

\bibitem{zhou2020sentix}
J.~Zhou, J.~Tian, R.~Wang, Y.~Wu, W.~Xiao, and L.~He, ``Sentix: A sentiment-aware pre-trained model for cross-domain sentiment analysis,'' in {\em Proceedings of the 28th international conference on computational linguistics}, pp.~568--579, 2020.

\bibitem{wang2020estimating}
P.~Wang, J.~Lai, Z.~Huang, Q.~Tan, and T.~Lin, ``Estimating traffic flow in large road networks based on multi-source traffic data,'' {\em IEEE Transactions on Intelligent Transportation Systems}, vol.~22, no.~9, pp.~5672--5683, 2020.

\bibitem{RAISSI2019686}
M.~Raissi, P.~Perdikaris, and G.~Karniadakis, ``Physics-informed neural networks: A deep learning framework for solving forward and inverse problems involving nonlinear partial differential equations,'' {\em Journal of Computational Physics}, vol.~378, pp.~686--707, 2019.

\bibitem{lv2014traffic}
Y.~Lv, Y.~Duan, W.~Kang, Z.~Li, and F.-Y. Wang, ``Traffic flow prediction with big data: A deep learning approach,'' {\em IEEE Transactions on Intelligent Transportation Systems}, vol.~16, no.~2, pp.~865--873, 2014.

\bibitem{raissi2018deep}
M.~Raissi, ``Deep hidden physics models: Deep learning of nonlinear partial differential equations,'' {\em The Journal of Machine Learning Research}, vol.~19, no.~1, pp.~932--955, 2018.

\bibitem{10105558}
A.~J. Huang and S.~Agarwal, ``On the limitations of physics-informed deep learning: Illustrations using first-order hyperbolic conservation law-based traffic flow models,'' {\em IEEE Open Journal of Intelligent Transportation Systems}, vol.~4, pp.~279--293, 2023.

\bibitem{park2016real}
H.~Park and A.~Haghani, ``Real-time prediction of secondary incident occurrences using vehicle probe data,'' {\em Transportation Research Part C: Emerging Technologies}, vol.~70, pp.~69--85, 2016.

\bibitem{greenshields1935study}
B.~D. Greenshields, J.~Bibbins, W.~Channing, and H.~Miller, ``A study of traffic capacity,'' in {\em Highway research board proceedings}, vol.~14, pp.~448--477, Washington, DC, 1935.

\bibitem{XING2022127079}
J.~Xing, W.~Wu, Q.~Cheng, and R.~Liu, ``Traffic state estimation of urban road networks by multi-source data fusion: Review and new insights,'' {\em Physica A: Statistical Mechanics and its Applications}, vol.~595, p.~127079, 2022.

\bibitem{9744160}
A.~K. Azad and M.~S. Islam, ``Traffic flow prediction model using google map and lstm deep learning,'' in {\em 2021 IEEE International Conference on Telecommunications and Photonics (ICTP)}, pp.~1--5, 2021.

\bibitem{8694956}
Z.~Zheng, Y.~Yang, J.~Liu, H.-N. Dai, and Y.~Zhang, ``Deep and embedded learning approach for traffic flow prediction in urban informatics,'' {\em IEEE Transactions on Intelligent Transportation Systems}, vol.~20, no.~10, pp.~3927--3939, 2019.

\end{thebibliography}

\end{document}